\documentclass[10pt,twocolumn,letterpaper]{article}

\usepackage{cvpr}              
\definecolor{cvprblue}{rgb}{0.21,0.49,0.74}
\usepackage[pagebackref,breaklinks,colorlinks,allcolors=cvprblue]{hyperref}

\usepackage{graphicx}
\usepackage{amsmath}
\usepackage{amssymb}
\usepackage{gensymb}

\usepackage{amssymb}
\usepackage{xcolor}

\usepackage{subcaption}
\usepackage{multirow}
\usepackage{algorithm}
\usepackage{algorithmic}

\usepackage{kotex}
\usepackage{color}
\usepackage{array}
\usepackage{hyperref}

\usepackage{bm}
\usepackage{nth}
\usepackage{setspace}

\newcolumntype{C}[1]{>{\centering\let\newline\\\arraybackslash\hspace{0pt}}m{#1}}
\newcolumntype{L}[1]{>{\let\newline\\\arraybackslash\hspace{0pt}}m{#1}}

\usepackage{rotating} 

\usepackage{amsmath}

\usepackage{listings}
\usepackage{xcolor}

\definecolor{codegray}{rgb}{0.5,0.5,0.5}
\definecolor{codepurple}{rgb}{0.58,0,0.82}
\definecolor{backcolor}{rgb}{0.98,0.98,0.98}

\lstdefinestyle{PythonStyle}{
    backgroundcolor=\color{backcolor},   
    commentstyle=\color{codegray},
    keywordstyle=\color{blue},
    numberstyle=\tiny\color{codegray},
    stringstyle=\color{codepurple},
    basicstyle=\ttfamily\footnotesize,
    breaklines=true,
    captionpos=b,                    
    keepspaces=true,                 
    numbers=left,                    
    numbersep=5pt,                  
    showspaces=false,                
    showstringspaces=false,
    showtabs=false,                  
    tabsize=2,
    language=Python                 
}
                        
\def\paperID{18029} 
\def\confName{CVPR}
\def\confYear{2026}

\title{DeFloMat: Detection with Flow Matching for \\ Stable and Efficient Generative Object Localization}

\author{
    Hansang Lee$^{1,}$\thanks{Equal contribution} \quad 
    Chaelin Lee$^{1,}$\footnotemark[1] \quad
    Nieun Seo$^2$ \quad 
    Joon Seok Lim$^2$ \quad 
    Helen Hong$^{1,}$\thanks{Corresponding author} \\
    $^1$Department of Software Convergence, Seoul Women's University, Seoul, Korea \\
    $^2$Department of Radiology, Severance Hospital, Yonsei University College of Medicine, Seoul, Korea \\
    {\tt\small \{hansanglee, cllee, hlhong\}@swu.ac.kr, \{sldmsdl, jslim1\}@yuhs.ac}
}

\begin{document}
\maketitle
\begin{abstract}
We propose DeFloMat (Detection with Flow Matching), a novel generative object detection framework that addresses the critical latency bottleneck of diffusion-based detectors, such as DiffusionDet, by integrating Conditional Flow Matching (CFM). Diffusion models achieve high accuracy by formulating detection as a multi-step stochastic denoising process, but their reliance on numerous sampling steps ($T \gg 60$) makes them impractical for time-sensitive clinical applications like Crohn's Disease detection in Magnetic Resonance Enterography (MRE). DeFloMat replaces this slow stochastic path with a highly direct, deterministic flow field derived from Conditional Optimal Transport (OT) theory, specifically approximating the Rectified Flow. This shift enables fast inference via a simple Ordinary Differential Equation (ODE) solver. We demonstrate the superiority of DeFloMat on a challenging MRE clinical dataset. Crucially, DeFloMat achieves state-of-the-art accuracy ($43.32\% \text{ } AP_{10:50}$) in only $3$ inference steps, which represents a $1.4\times$ performance improvement over DiffusionDet's maximum converged performance ($31.03\% \text{ } AP_{10:50}$ at $4$ steps). Furthermore, our deterministic flow significantly enhances localization characteristics, yielding superior Recall and stability in the few-step regime. DeFloMat resolves the trade-off between generative accuracy and clinical efficiency, setting a new standard for stable and rapid object localization.
\end{abstract}    
\begin{figure}[!]
    \centering
    \begin{subfigure}[b]{\linewidth}
        \centering
        \includegraphics[width=\linewidth]{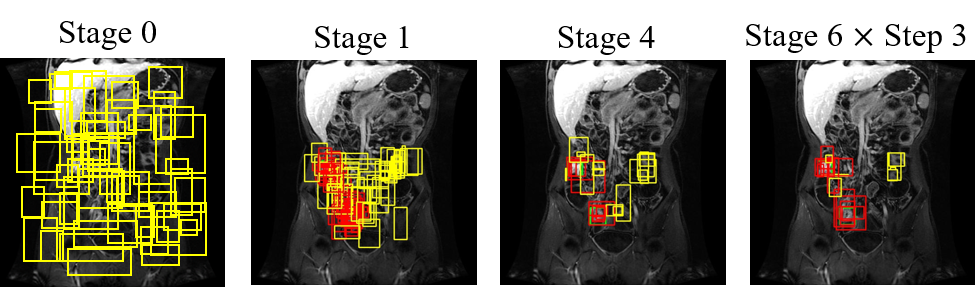}
        \caption{DiffusionDet}
        \label{fig:vertical:a}
    \end{subfigure}
    \begin{subfigure}[b]{\linewidth}
        \centering
        \includegraphics[width=\linewidth]{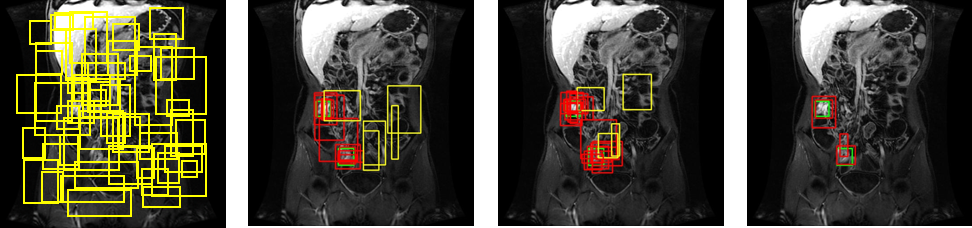}
        \caption{DeFloMat}
        \label{fig:vertical:b}
    \end{subfigure}
    \begin{subfigure}[b]{\linewidth}
        \centering
        \includegraphics[width=\linewidth]{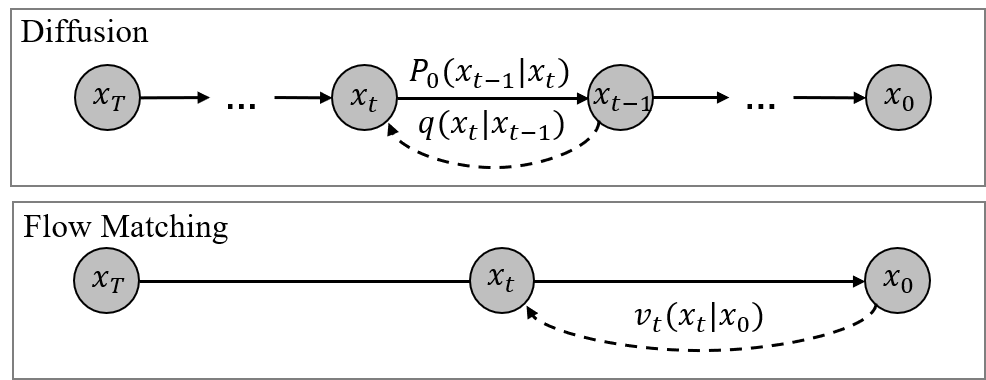}
        \caption{Diffusion vs. Flow Matching}
        \label{fig:vertical:c}
    \end{subfigure}
    \caption{\textbf{Comparison of generative localization dynamics between DiffusionDet and DeFloMat.} 
    (a) \textbf{DiffusionDet} requires many stochastic denoising stages to stabilize bounding boxes, often producing highly scattered proposals in early steps.
    (b) Our \textbf{DeFloMat} rapidly converges to anatomically plausible inflammation regions within only a few deterministic ODE steps, yielding sharper and more stable localization.
    (c) \textbf{Diffusion vs. Flow Matching.} Diffusion performs iterative stochastic reverse transitions ($p_\theta(x_{t-1}\mid x_t)$), whereas DeFloMat learns a direct optimal-transport vector field ($v_t(x_t\mid x_0)$) that enables fast, stable integration from noise to final boxes.}
    \label{fig:teaser}
\end{figure}

\section{Introduction}
\label{sec:intro}

Accurate and timely diagnosis of inflammatory activity in Inflammatory Bowel Disease (IBD), particularly Crohn's Disease (CD), is essential for effective patient management and therapeutic response assessment~\cite{ecco2017guidelines}. Magnetic Resonance Enterography (MRE) has emerged as a critical non-invasive modality, offering comprehensive visualization of both luminal and extraluminal bowel pathology, making it ideal for longitudinal monitoring without the risks associated with ionizing radiation~\cite{Taylor2018METRIC,Tsai2019MRE}. Despite the utility of MRE, the current gold standard for assessing disease activity relies heavily on manual interpretation using semi-quantitative scoring systems such as the Magnetic Resonance Index of Activity (MaRIA)~\cite{rimola2011maria}. These methods are prone to high inter-observer variability and require interpretation by highly specialized radiologists, leading to inconsistencies and bottlenecks in clinical workflow. The development of deep learning models capable of objectively and rapidly detecting colonic inflammatory activity is therefore paramount to standardizing diagnosis and building a reliable computer-aided diagnostic (CAD) system that can ensure timely intervention and therapy adjustment~\cite{Tsai2019MRE}. The inherent challenges of this task—subtle visual cues, variable inflammation severity, and high class imbalance—necessitate robust and highly precise object localization techniques.

Traditional object detection methods, such as anchor-based (e.g., Faster R-CNN~\cite{girshick2015fast}, YOLOv4~\cite{Bochkovskiy2020YOLOv4}) and anchor-free models (e.g., Sparse R-CNN~\cite{sun2021sparsercnn}, DETR~\cite{carion2020detr}), have achieved remarkable success in general computer vision tasks. However, in complex medical imaging scenarios like MRE, they often struggle with the precise localization of subtle pathological findings due to their limited ability to model the full posterior distribution of the bounding boxes. Recent works have explored the powerful generative capabilities of Diffusion Models (DMs)~\cite{ho2020ddpm} to address this limitation. DiffusionDet~\cite{Chen2023DiffusionDet_ICCV} is a notable example that formulates object detection as a generative process, where the ground-truth bounding box proposals $\mathbf{x}_0$ are gradually diffused to noise $\mathbf{x}_T$ through a Markovian chain of small stochastic steps $q(\mathbf{x}_t|\mathbf{x}_{t-1})$. The model then learns the reverse process $p_{\theta}(\mathbf{x}_{t-1}|\mathbf{x}_t)$ to recover the original boxes via iterative denoising. While DiffusionDet provides superior accuracy and localization stability compared to its deterministic counterparts, it suffers from a critical drawback: latency. High-quality detection typically requires a large number of inference steps (e.g., $T \gg 60$). This limitation necessitates a new paradigm that retains the generative power of DMs while achieving few-step, efficient inference.

In this paper, we propose DeFloMat (Detection with Flow Matching), a novel generative object detection framework that overcomes the speed-accuracy trade-off inherent in diffusion-based detectors. DeFloMat is built upon the insight that the slow, multi-step stochastic diffusion process can be replaced by a single, direct, deterministic flow field. Specifically, we leverage Conditional Flow Matching (CFM)~\cite{liu2022flow,Lipman2022FlowMatching} to learn the optimal transport (OT) path between the prior noise distribution $\mathbf{x}_T$ and the target data distribution $\mathbf{x}_0$. This path is approximated by the simple Rectified Flow, and our detection decoder is trained to predict the velocity vector field $\mathbf{v}_t$ that governs the flow. This fundamental shift allows us to transform the slow ancestral sampling into a rapid inference via a standard Ordinary Differential Equation (ODE) solver. 

Our key contributions are summarized as follows: 
\begin{itemize}
    \item We introduce the first generative object detector, DeFloMat, based on Conditional Flow Matching, achieving unprecedented efficiency in the generative detection space. 
    \item We demonstrate that DeFloMat significantly outperforms DiffusionDet on a challenging MRE clinical dataset, achieving $43.32\% \text{ } AP_{10:50}$ in only 3 inference steps, representing a $\mathbf{1.4\times}$ performance improvement over DiffusionDet's maximum converged performance ($31.03\% \text{ } AP_{10:50}$ at 4 steps). 
    \item The deterministic nature of the flow matching model is shown to dramatically improve the localization characteristics and stability in the few-step inference regime, making it highly suitable for reliable CAD systems.
\end{itemize}
\begin{figure*}[t!]
    \centering
    \includegraphics[width=0.95\linewidth]{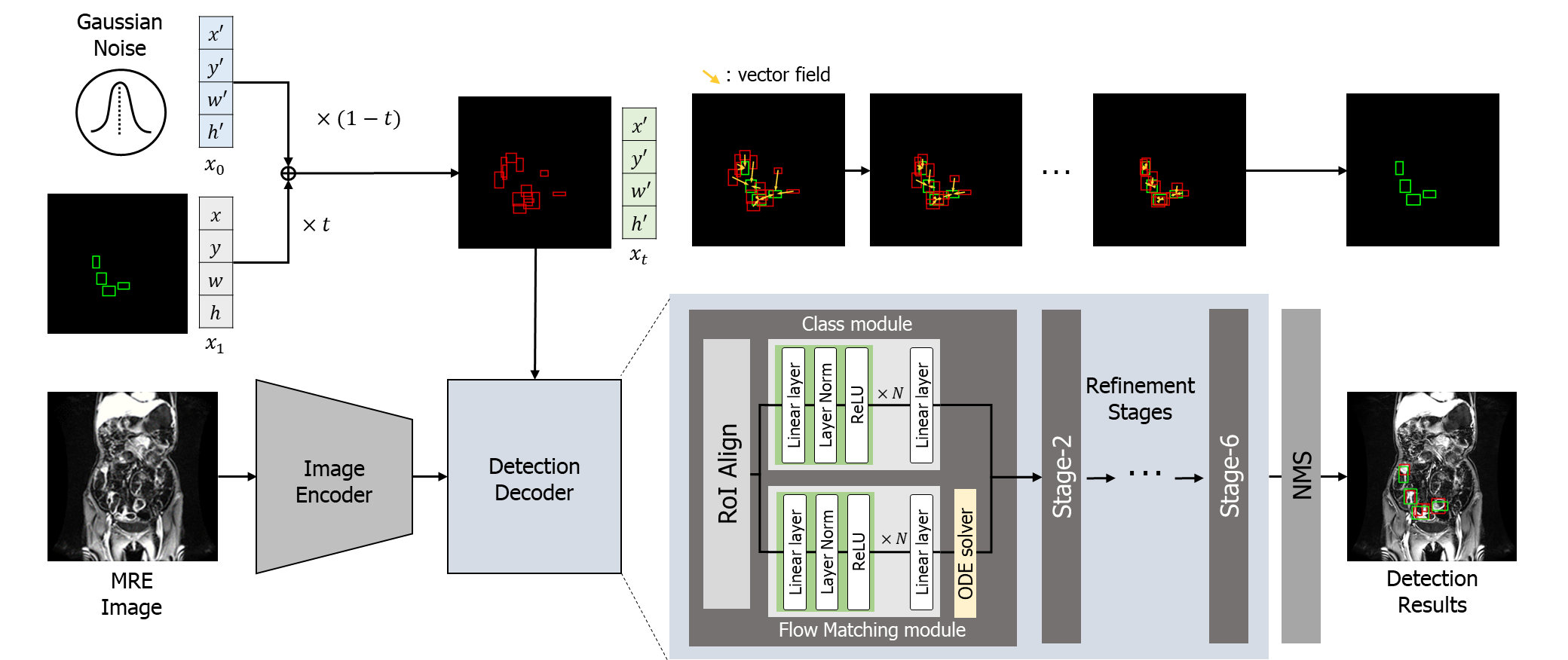}
    \caption{\textbf{The Overall Architecture of DeFloMat.}
    The DeFloMat pipeline consists of an Image Encoder (ResNet50), a Detection Decoder, and a novel \textbf{Flow Matching Module}. During training, the decoder is conditioned on MRE features ($F$) and time ($t$) to predict the deterministic velocity vector field ($\mathbf{v}_{pred}$) that transports the noisy box proposals ($\mathbf{x}_t$) directly towards the ground truth ($\mathbf{x}_1$) along a straight Optimal Transport path ($\mathbf{x}_t = (1-t) \cdot \mathbf{x}_0 + t \cdot \mathbf{x}_1$). The network is supervised by a specialized \textbf{flow matching loss} $\mathcal{L}_{\text{flow}}$. Crucially, during inference, the deterministic flow enables us to replace the slow stochastic diffusion sampling with a single ODE Solver (e.g., Euler update) integrated within the refinement stages (Stage 2-6), dramatically accelerating object localization with minimal steps ($S \le 4$). The final results are obtained after Non-Maximum Suppression (NMS).}
    \label{fig:method}
\end{figure*}

\section{Methods}
\label{sec:method}

\subsection{Architecture}

The DeFloMat framework integrates Conditional Flow Matching (CFM) into the generative detection paradigm pioneered by DiffusionDet, achieving superior efficiency and stability. As illustrated in Figure~\ref{fig:method}, our pipeline is composed of three main interlinked modules: the Image Encoder, the Detection Decoder, and the novel Flow Matching Module. The Image Encoder extracts features from the MRE image slice, which serve as the conditioning context $F$. The Detection Decoder then takes noisy box proposals $\mathbf{x}_t$, the feature $F$, and the current continuous timestep $t$ to predict the instantaneous velocity vector field $\mathbf{v}_{pred}$. During training, the Flow Matching Module supervises the learning of this deterministic field. Crucially, during inference, the deterministic nature of the learned flow allows us to leverage an Ordinary Differential Equation (ODE) solver for rapid, few-step sampling, completely eliminating the slow stochasticity of traditional diffusion sampling.

\paragraph{Image Encoder.}
Consistent with state-of-the-art detection models, the Image Encoder is responsible for extracting rich, multi-scale features from the MRE input. We employ a ResNet50~\cite{he2016resnet} backbone initialized with weights pre-trained on ImageNet-1K~\cite{deng2009imagenet}, followed by a Feature Pyramid Network (FPN)~\cite{lin2017fpn} structure. The encoder runs only once per MRE slice, generating a deep feature representation $F$ that conditions the subsequent box refinement process. The primary function of $F$ is to provide the necessary spatial and semantic context for the Detection Decoder to accurately predict the flow vector field required for object localization.

\paragraph{Detection Decoder with Flow Matching Module.}
The Detection Decoder is the core generative component, responsible for iteratively transforming noisy box proposals $\mathbf{x}_t$ into final bounding box predictions $\mathbf{x}_1$. It is a shared-weight, multi-stage structure (six cascading stages, similar to transformer-based detectors) that takes three inputs: the current box state $\mathbf{x}_t$, the image features $F$, and the continuous timestep embedding $t$. Unlike DiffusionDet, which predicts a denoised state or a small refinement offset, our decoder predicts the instantaneous velocity vector field $\mathbf{v}_{pred} = \mathbf{v}_{\theta}(\mathbf{x}_t, F, t)$. This vector represents the required movement in the bounding box coordinate space to move along the deterministic flow path.

The Flow Matching Module defines the learning objective for this prediction. It constructs the target straight-line path (Rectified Flow) between the initial noise $\mathbf{x}_0$ and the ground truth $\mathbf{x}_1$ as $\mathbf{x}_t = (1-t) \cdot \mathbf{x}_0 + t \cdot \mathbf{x}_1$. The target velocity field $\mathbf{u}_t$ is simply the constant difference: $\mathbf{u}_t = \mathbf{x}_1 - \mathbf{x}_0$. The decoder is then trained to minimize the $\ell_2$ distance between its prediction and this target flow, which is detailed in the subsequent section on the loss function. This dual-role architecture—feature extraction followed by flow-based, iterative refinement—is key to DeFloMat's high efficiency and accuracy.

\subsection{Training}

The training of DeFloMat is designed to enforce the prediction of the deterministic velocity vector field, distinguishing it significantly from the stochastic noise prediction in DiffusionDet. The training process involves defining the ground-truth boxes ($\mathbf{x}_1$), corrupting them via the straight-line path, and optimizing a composite loss function that includes the Flow Matching regularization.

\paragraph{Groud-Truth Box Preparation.}
Since the number of ground-truth (GT) boxes varies per MRE slice, we align the box set for training by adopting the standard practice of padding. We define a fixed maximum number of proposals, $N_{prop}$ (e.g., 12 in training). The actual GT boxes are repeated and/or padded with dummy boxes until the total count reaches $N_{prop}$. Each box coordinate $\mathbf{x} = (x, y, w, h)$ is normalized to the $[0, 1]$ range relative to the image size. For enhanced flow stability, especially concerning scale parameters ($w, h$), which are susceptible to flow divergence, we perform the entire flow operation in the logarithmic space for scale. The GT boxes are transformed as $\mathbf{x}_1 = (\text{norm}(x), \text{norm}(y), \log(w), \log(h))$.

\paragraph{Box Corruption and Flow Target Definition.}
The core of our approach is the construction of the non-Markovian, deterministic path derived from Rectified Flow. We first sample the initial noise $\mathbf{x}_0$ (from $\mathcal{U}(0, 1)$ or $\mathcal{N}(0, 1)$ in the logarithmic space) and a continuous time step $t \sim \mathcal{U}(0, 1)$. The corrupted box state $\mathbf{x}_t$ is defined by the linear interpolation between the noise and the ground truth:
\begin{equation}
\mathbf{x}_t = (1-t) \cdot \mathbf{x}_0 + t \cdot \mathbf{x}_1
\end{equation}

This path represents the most efficient, non-curving probability path between the endpoints in the box space. The target velocity vector field $\mathbf{u}_t$ is the constant velocity required to move instantaneously along this straight path:

\begin{equation}
\mathbf{u}_t = \frac{d\mathbf{x}_t}{dt} = \mathbf{x}_1 - \mathbf{x}_0
\end{equation}

The Detection Decoder $\mathbf{v}_{\theta}(\mathbf{x}_t, F, t)$ is trained to approximate this true target flow $\mathbf{u}_t$, conditioned on the image feature $F$ and time $t$.

\paragraph{Loss Function.}
DeFloMat optimizes a multi-task loss function that balances standard object detection objectives with the novel Flow Matching regularization term:

\begin{equation}
\mathcal{L}_{\text{total}} = \mathcal{L}_{\text{cls}} + \lambda_{L1} \cdot \mathcal{L}_{L1} + \lambda_{\text{GloU}} \cdot \mathcal{L}_{\text{GloU}} + \lambda_{\text{flow}} \cdot \mathcal{L}_{\text{flow}}
\end{equation}

where $\mathcal{L}_{\text{cls}}$ is the Focal Loss for classification, $\mathcal{L}_{L1}$ and $\mathcal{L}_{\text{GloU}}$~\cite{rezatofighi2019giou} are the standard box regression losses (Smooth L1 and Generalized IoU) for the predicted final box $\mathbf{\tilde{x}}_1$, and $\lambda$ terms are the loss weights.

The key addition is the Flow Matching Loss ($\mathcal{L}_{\text{flow}}$), which explicitly enforces the deterministic flow constraint by minimizing the $\ell_2$ distance between the predicted flow $\mathbf{v}_{\theta}$ and the target flow $\mathbf{u}_t$:

\begin{equation}
\mathcal{L}_{\text{flow}} = \| \mathbf{v}_{\theta}(\mathbf{x}_t, F, t) - \mathbf{u}_t \|_2^2
\end{equation}

The predicted final box $\mathbf{\tilde{x}}_1$ used for the detection losses is obtained by integrating the predicted flow from $\mathbf{x}_t$ to $\mathbf{x}_1$: $\mathbf{\tilde{x}}_1 = \mathbf{x}_t + (1-t) \cdot \mathbf{v}_{\theta}(\mathbf{x}_t, F, t)$. The optimal weights determined in our experiments were $\lambda_{L1}=2.0$, $\lambda_{\text{GloU}}=5.0$, $\lambda_{\text{cls}}=2.0$, and crucially, a light regularization weight $\lambda_{\text{flow}}=0.1$.
The training procedure is summarized in Algorithm~\ref{alg:train}.

\begin{algorithm}[t]
\caption{Training Loss Computation for DeFloMat}
\label{alg:train}
\begin{algorithmic}[1]

\REQUIRE Images $\mathbf{I} \in \mathbb{R}^{B\times H\times W\times 3}$, 
GT boxes $\mathbf{B} \in \mathbb{R}^{B\times *\times 4}$
\ENSURE Total loss $L_{\text{total}}$

\STATE $\mathbf{F} \leftarrow \text{ImageEncoder}(\mathbf{I})$

\STATE $\mathbf{x}_1 \leftarrow \text{PadBoxes}(\mathbf{B})$ \hfill // $[B,N,4]$
\STATE $\mathbf{x}_1 \leftarrow \log(\mathbf{x}_1)$

\STATE $\mathbf{x}_0 \sim \text{Uniform}(0,1)$ \hfill // $[B,N,4]$
\STATE $\mathbf{x}_0 \leftarrow \log(\mathbf{x}_0)$
\STATE $t \sim \mathcal{U}(0,1)$

\STATE $\mathbf{x}_t = (1-t)\mathbf{x}_0 + t\mathbf{x}_1$
\STATE $\mathbf{u}_t = \mathbf{x}_1 - \mathbf{x}_0$

\STATE $\mathbf{v}_{\text{pred}} = 
\text{DetectionDecoder}(\exp(\mathbf{x}_t), \mathbf{F}, t)$

\STATE $\mathbf{p}_t = \mathbf{x}_t + (1-t)\mathbf{v}_{\text{pred}}$
\STATE $\mathbf{p}_t = \exp(\mathbf{p}_t)$

\STATE $(L_{\text{cls}}, L_{L1}, L_{\text{GIoU}}, L_{\text{flow}})
       \leftarrow \text{SetPredictionLoss}(\mathbf{p}_t, \mathbf{B}, 
       \mathbf{v}_{\text{pred}}, \mathbf{u}_t)$

\STATE $L_{\text{total}} = L_{\text{cls}} + L_{L1} + L_{\text{GIoU}} + L_{\text{flow}}$

\RETURN $L_{\text{total}}$

\end{algorithmic}
\end{algorithm}

\subsection{Inference}

The primary motivation for DeFloMat is to overcome the high latency of generative detection. The learned deterministic velocity field $\mathbf{v}_{\theta}$ (Eq. 2) allows us to replace the slow stochastic sampling of DiffusionDet with a rapid, direct integration of an Ordinary Differential Equation (ODE).

\paragraph{Sampling Steps.}
Since $\mathbf{v}_{\theta}$ represents the direct Optimal Transport velocity from noise $\mathbf{x}_0$ to data $\mathbf{x}_1$, the transformation can be accurately approximated with a simple Euler Update using very few steps $S$. At inference, a large set of initial proposals $\mathbf{p}_0$ (e.g., $N_{prop}=120$) is sampled from $\mathcal{N}(0, 1)$ in the latent space. The Euler integration iteratively refines these proposals:

\begin{equation}
\mathbf{x}_{t+\Delta t} = \mathbf{x}_t + \Delta t \cdot \mathbf{v}_{\theta}(\mathbf{x}_t, \mathbf{F}, t)
\end{equation}

where $\Delta t = 1/S$. Our empirical results confirm that DeFloMat achieves maximum or near-maximum accuracy at $S=3$ steps, resulting in an unprecedented efficiency gain compared to the minimum $T \gg 60$ steps typically required by the DDIM~\cite{song2021ddim} path of DiffusionDet.

\paragraph{Elimination of Box Renewal.}
The deterministic and stable nature of the learned flow field inherently provides robust performance in the few-step regime. This stability removes the necessity for heuristic post-processing techniques that were vital for DiffusionDet to stabilize its stochastic predictions. Specifically, DeFloMat's inference pipeline does not require the use of: (1) Box Renewal, which is used to replace low-confidence boxes with new noise mid-process, or (2) Ensemble methods, which average predictions from multiple diffusion streams. The final result is obtained by simply transforming the log-space outputs $\mathbf{p}_t$ back to coordinates and applying a standard Non-Maximum Suppression (NMS)~\cite{bodla2017softnms}.
The inference procedure is summarized in Algorithm~\ref{alg:infer}.

\begin{algorithm}[t]
\caption{DeFloMat Inference via ODE Integration}
\label{alg:infer}
\begin{algorithmic}[1]

\REQUIRE Images $\mathbf{I}$, number of steps $S$
\ENSURE Final predicted boxes $\mathbf{p}$

\STATE $\mathbf{F} \leftarrow \text{ImageEncoder}(\mathbf{I})$

\STATE $\mathbf{p}_t \sim \mathcal{N}(0,1)$ \hfill // initial noisy boxes

\STATE $\{t_0,\dots,t_S\} \leftarrow \text{linspace}(1,0,S)$
\STATE $\mathcal{T} \leftarrow \text{zip}(t_0,\dots,t_{S-1};\, t_1,\dots,t_S)$

\FOR{each $(t_{\text{now}}, t_{\text{next}})$ in $\mathcal{T}$}
    \STATE $\Delta t = t_{\text{next}} - t_{\text{now}}$
    \STATE $\mathbf{f} \leftarrow 
        \text{DetectionDecoder}(\mathbf{p}_t, \mathbf{F}, t_{\text{now}})$
    \STATE $\mathbf{p}_t \leftarrow \mathbf{p}_t + \Delta t \cdot \mathbf{f}$ 
          \hfill // Euler update
\ENDFOR

\RETURN $\mathbf{p}_t$

\end{algorithmic}
\end{algorithm}
\begin{table*}[t]
\centering
\small
\setlength{\tabcolsep}{5pt}  
\caption{\textbf{Performance Comparison on Crohn's Disease MRE Detection.} Evaluation results are presented for different confidence thresholds ($\tau_c$) using 12 proposals ($N_{prop}=12$) and only $S=3$ inference steps for the generative models (DiffusionDet, DeFloMat). DeFloMat demonstrates state-of-the-art performance, achieving a significantly higher sensitivity ($\text{Recall}_{10}$) and localization accuracy ($\text{AP}_{10}$) compared to DiffusionDet and conventional methods (YOLOV4, Mask R-CNN). The gains highlight the effectiveness of flow matching in stabilizing generative detection in the highly efficient, few-step regime.}
\label{tab:performance_comparison}

\begin{tabular}{cc|cc|cc|cc}
\toprule
\textbf{$\tau_{c}$ (\%)} & \textbf{Methods} 
& \textbf{AP$_{10}$} & \textbf{AP$_{50}$}
& \textbf{Precision$_{10}$} & \textbf{Precision$_{50}$}
& \textbf{Recall$_{10}$} & \textbf{Recall$_{50}$} \\
\midrule

\multirow{5}{*}{30}
& YOLOv4~\cite{Bochkovskiy2020YOLOv4} 
& 20.04 & 14.66 & 29.13 & 21.26 & 20.27 & 14.98 \\

& Mask R-CNN~\cite{He2017MaskRCNN} 
& 40.63 & 22.05 & 40.41 & 22.00 & 44.74 & 24.78 \\

& DiffusionDet~\cite{Chen2023DiffusionDet_ICCV}
& 49.13 & \textbf{31.51} & \textbf{53.11} & \textbf{34.05} & 51.75 & 33.76 \\


& \textbf{DeFloMat (Ours)} 
& \textbf{54.37} & 31.29 & 48.15 & 29.37 & \textbf{59.17} & \textbf{34.31} \\
\midrule

\multirow{5}{*}{40}
& YOLOv4~\cite{Bochkovskiy2020YOLOv4} 
& 14.04 & 11.24 & 21.85 & 17.36 & 14.14 & 11.81 \\

& Mask R-CNN~\cite{He2017MaskRCNN} 
& 35.15 & 20.09 & 38.89 & 21.97 & 37.48 & 21.38 \\

& DiffusionDet~\cite{Chen2023DiffusionDet_ICCV}
& 41.78 & 27.79 & 53.07 & \textbf{35.59} & 43.33 & 29.24 \\


& \textbf{DeFloMat (Ours)} 
& \textbf{53.27} & \textbf{31.17} & \textbf{56.47} & 34.24 & \textbf{57.06} & \textbf{34.05} \\
\midrule

\multirow{5}{*}{50}
& YOLOv4~\cite{Bochkovskiy2020YOLOv4} 
& 9.46 & 7.69 & 15.52 & 12.42 & 9.53 & 7.77 \\

& Mask R-CNN~\cite{He2017MaskRCNN} 
& 30.92 & 18.28 & 37.28 & 21.72 & 32.39 & 19.48 \\

& DiffusionDet~\cite{Chen2023DiffusionDet_ICCV}
& 34.32 & 24.08 & 46.87 & 32.95 & 35.21 & 25.02 \\


& \textbf{DeFloMat (Ours)} 
& \textbf{52.63} & \textbf{31.16} & \textbf{58.22} & \textbf{35.53} & \textbf{56.12} & \textbf{34.01} \\
\bottomrule

\end{tabular}
\end{table*}

\begin{figure}[t]
    \centering
    \begin{subfigure}[b]{0.3\linewidth}
        \centering
        \includegraphics[width=\textwidth]{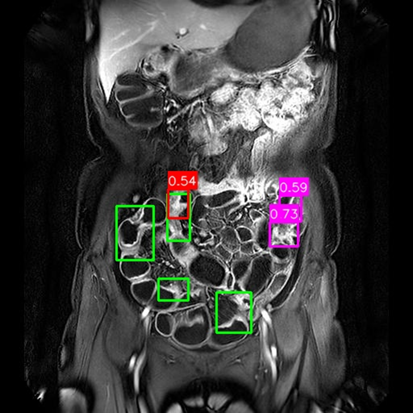}
    \end{subfigure}
    \begin{subfigure}[b]{0.3\linewidth}
        \centering
        \includegraphics[width=\textwidth]{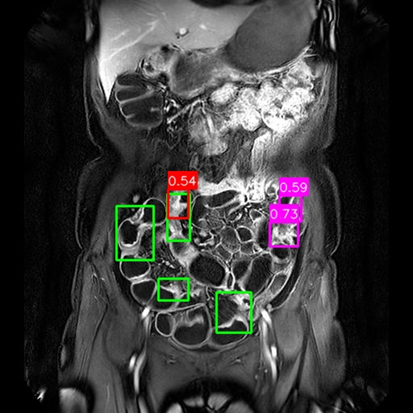}
    \end{subfigure}
    \begin{subfigure}[b]{0.3\linewidth}
        \centering
        \includegraphics[width=\textwidth]{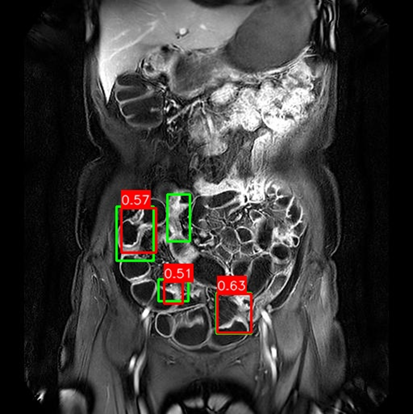}
    \end{subfigure}
    \begin{subfigure}[b]{0.3\linewidth}
        \centering
        \includegraphics[width=\textwidth]{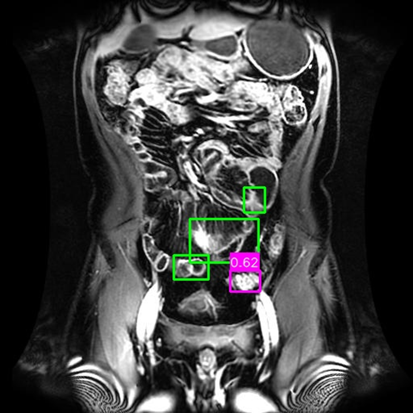}
    \end{subfigure}
    \begin{subfigure}[b]{0.3\linewidth}
        \centering
        \includegraphics[width=\textwidth]{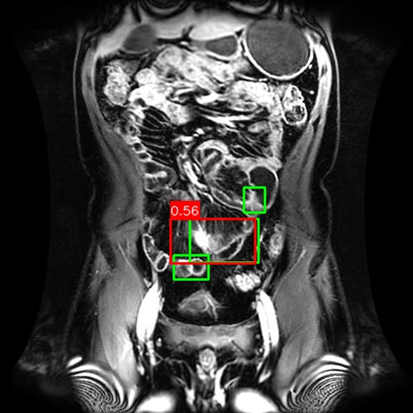}
    \end{subfigure}
    \begin{subfigure}[b]{0.3\linewidth}
        \centering
        \includegraphics[width=\textwidth]{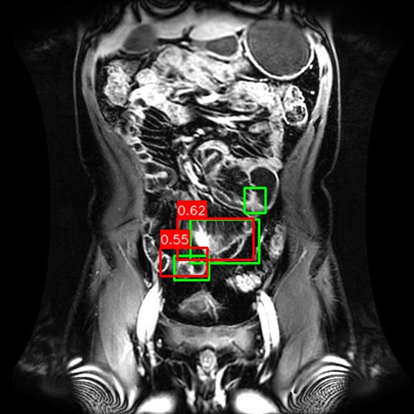}
    \end{subfigure}
    \begin{subfigure}[b]{0.3\linewidth}
        \centering
        \includegraphics[width=\textwidth]{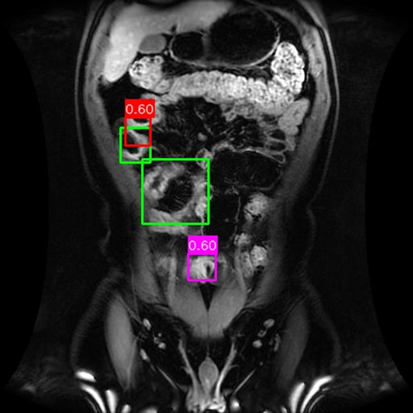}
    \end{subfigure}
    \begin{subfigure}[b]{0.3\linewidth}
        \centering
        \includegraphics[width=\textwidth]{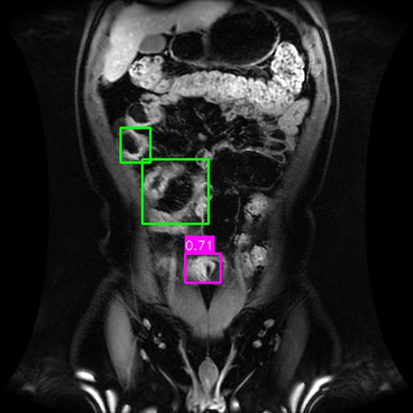}
    \end{subfigure}
    \begin{subfigure}[b]{0.3\linewidth}
        \centering
        \includegraphics[width=\textwidth]{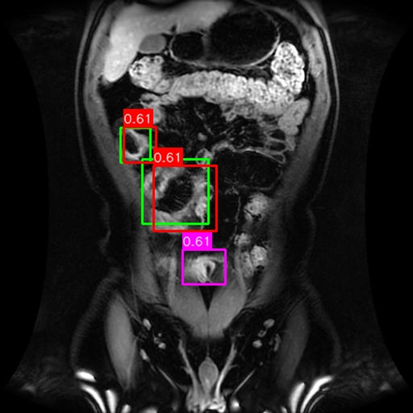}
    \end{subfigure}
    \begin{subfigure}[b]{0.3\linewidth}
        \centering
        \includegraphics[width=\textwidth]{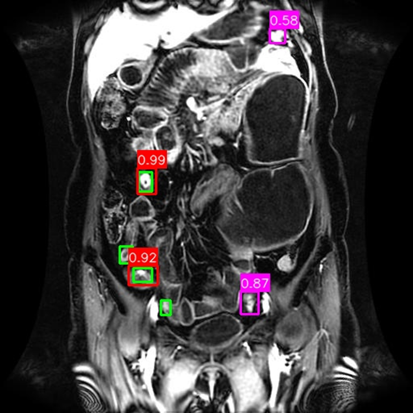}
    \end{subfigure}
    \begin{subfigure}[b]{0.3\linewidth}
        \centering
        \includegraphics[width=\textwidth]{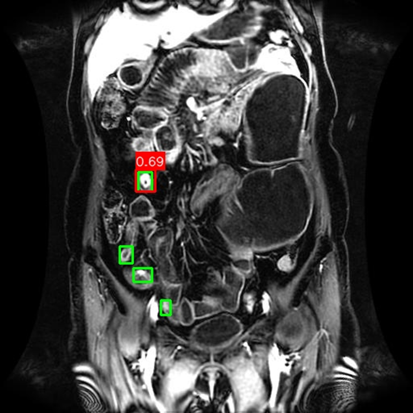}
    \end{subfigure}
    \begin{subfigure}[b]{0.3\linewidth}
        \centering
        \includegraphics[width=\textwidth]{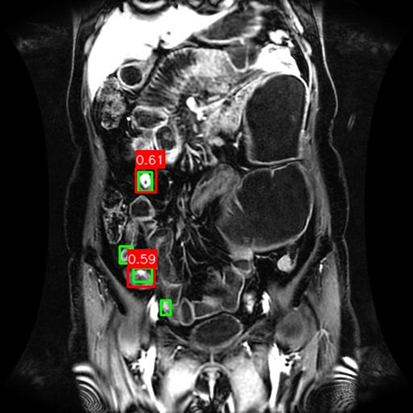}
    \end{subfigure}
    \begin{subfigure}[b]{0.3\linewidth}
        \centering
        \includegraphics[width=\textwidth]{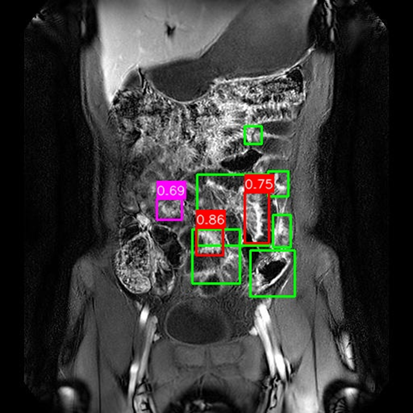}
    \end{subfigure}
    \begin{subfigure}[b]{0.3\linewidth}
        \centering
        \includegraphics[width=\textwidth]{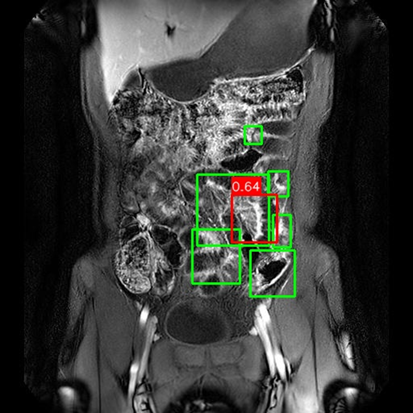}
    \end{subfigure}
    \begin{subfigure}[b]{0.3\linewidth}
        \centering
        \includegraphics[width=\textwidth]{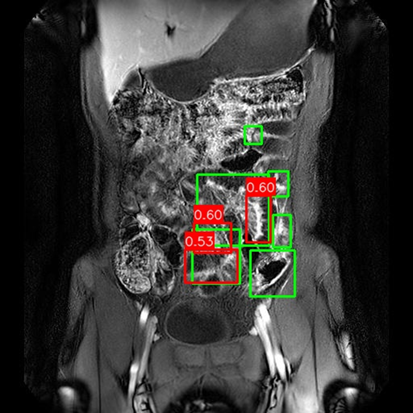}
    \end{subfigure}
    \begin{subfigure}[b]{0.3\linewidth}
        \centering
        \includegraphics[width=\textwidth]{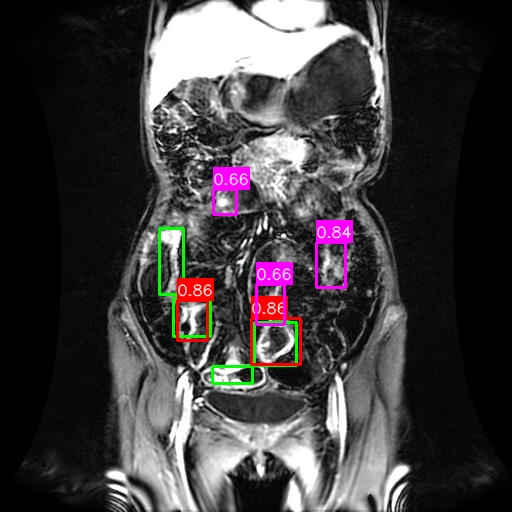}
        \caption{Mask R-CNN~\cite{He2017MaskRCNN}}
        \label{fig:ex_a}
    \end{subfigure}
    \begin{subfigure}[b]{0.3\linewidth}
        \centering
        \includegraphics[width=\textwidth]{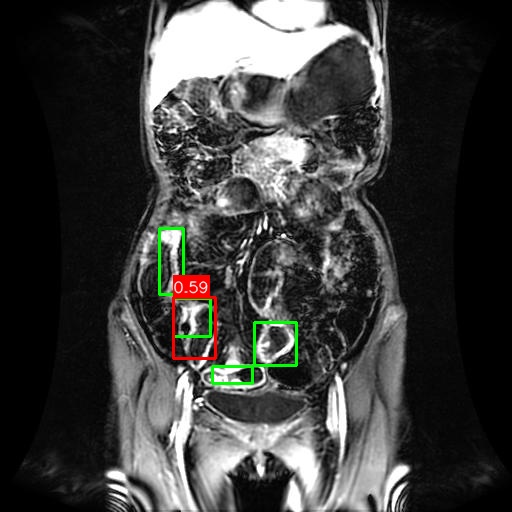}
        \caption{DiffusionDet~\cite{Chen2023DiffusionDet_ICCV}}
        \label{fig:ex_b}
    \end{subfigure}
    \begin{subfigure}[b]{0.3\linewidth}
        \centering
        \includegraphics[width=\textwidth]{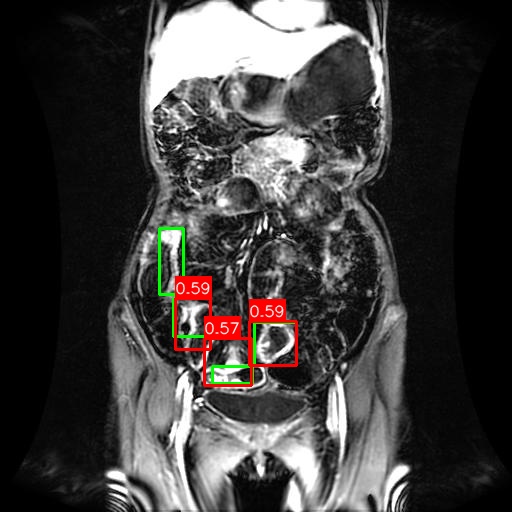}
        \caption{DeFloMat (Ours)}
        \label{fig:ex_c}
    \end{subfigure}
    \caption{\textbf{Qualitative Comparison on Crohn's Disease MRE Test Set.} The figure compares detection results from (a) Mask R-CNN, (b) DiffusionDet ($S=3$), and (c) DeFloMat (Ours, $S=3$) on challenging MRE slices. Green boxes indicate True Positives (TP, IoU $\ge 0.1$), Red boxes indicate False Positives (FP), and Purple boxes indicate False Negatives (FN). DeFloMat consistently demonstrates superior localization quality and sensitivity: it successfully detects subtle inflammation regions (TP) that are often missed (FN, Purple boxes) by the Mask R-CNN baseline (Row 2, 4). Furthermore, DeFloMat provides tighter bounding box localization compared to DiffusionDet, confirming the benefit of learning the direct, deterministic flow field. The results show DeFloMat's robustness in capturing varying sizes and numbers of inflammatory lesions.}
    \label{fig:examples}
\end{figure}

\begin{figure}[t]
    \centering
    \begin{subfigure}[b]{\linewidth}
        \centering
        \includegraphics[width=\textwidth]{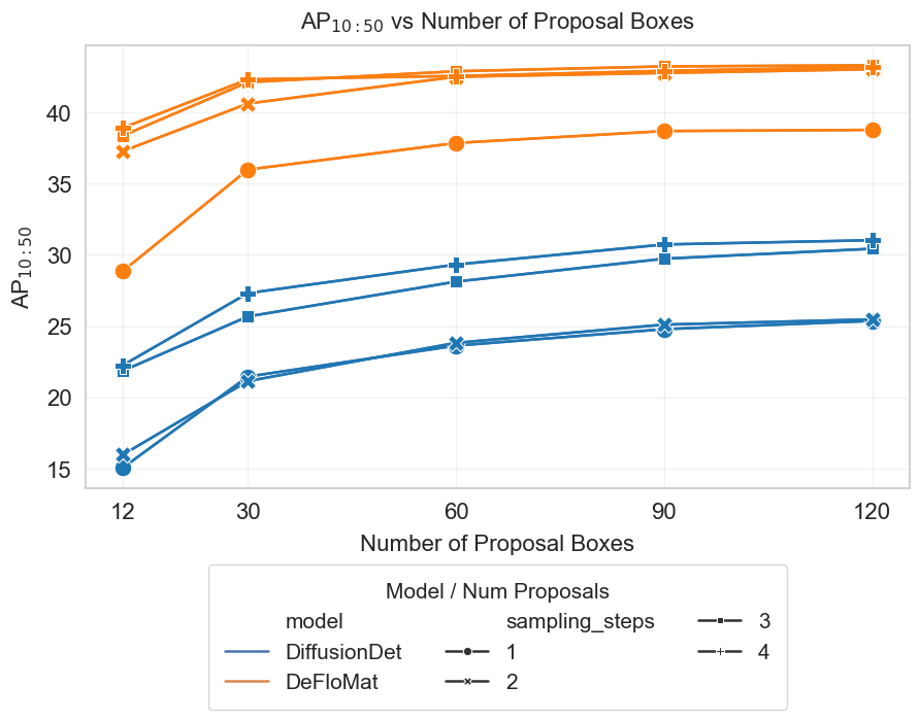}
        \caption{Number of Box Proposals vs. AP$_{10:50}$}
        \label{fig:box_vs_ap}
    \end{subfigure}
    \begin{subfigure}[b]{\linewidth}
        \centering
        \includegraphics[width=\textwidth]{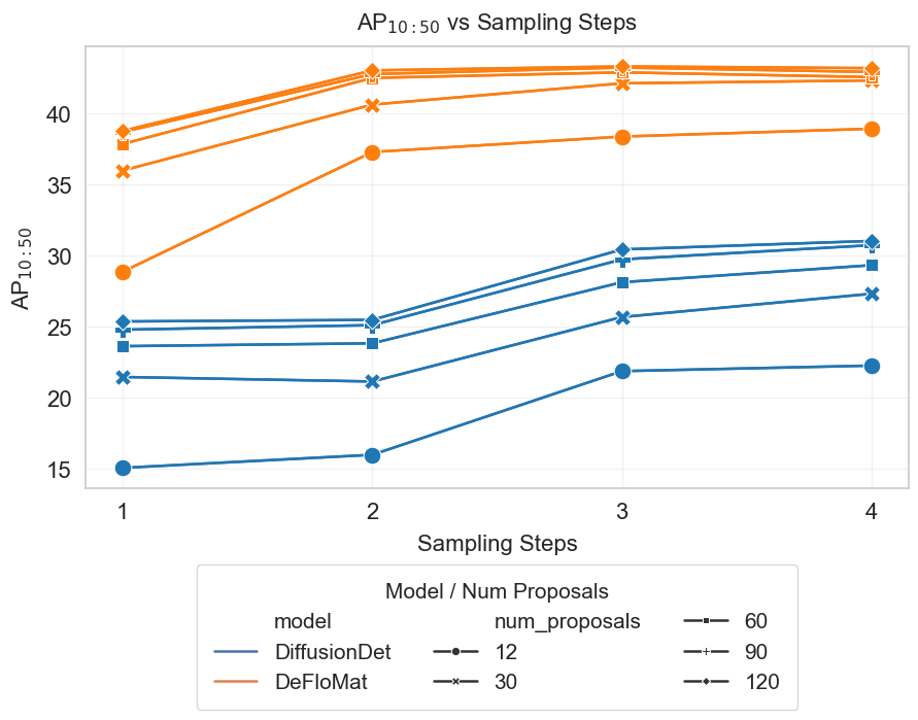}
        \caption{Sampling Steps vs. AP$_{10:50}$}
        \label{fig:step_vs_ap}
    \end{subfigure}
    \caption{\textbf{Efficiency and Stability Analysis of DeFloMat vs. DiffusionDet} 
    (a) Number of Proposal Boxes ($N_{prop}$) vs. $AP_{10:50}$: DeFloMat (Ours, orange) achieves consistently higher $AP_{10:50}$ than DiffusionDet (blue) and maintains stable performance across increasing $N_{prop}$, demonstrating robustness even in the low-step regime ($S \le 4$).
    (b) Sampling Steps ($S$) vs. $AP_{10:50}$: DeFloMat exhibits superior convergence efficiency, saturating at $S=3$ steps with an accuracy $\mathbf{1.4\times}$ higher than DiffusionDet's peak performance. This validates the deterministic flow path's efficacy in achieving high accuracy with minimal computational cost.}
    \label{fig:vs}
\end{figure}

\section{Experiments}
\label{sec:experiment}

\subsection{Experimental Settings}

\paragraph{Dataset.}
We evaluate DeFloMat on a proprietary clinical dataset focusing on the detection of inflammatory activity in Crohn's Disease (CD) using Magnetic Resonance Enterography (MRE) images. The dataset comprises 208 CD patients from a major hospital, with $180$ patients showing active inflammation. All active sites were meticulously annotated with bounding boxes by clinical experts. The dataset was split based on patient IDs to ensure no patient cross-contamination between sets: $\mathbf{104}$ patients for training (4,177 boxes), $\mathbf{52}$ for validation (1,766 boxes), and $\mathbf{52}$ for testing (2,165 boxes). Input MRE images were resampled and standardized to $\mathbf{512 \times 512}$ resolution.

\paragraph{Implementation Details.}
Our models are built upon the PyTorch framework using the Detectron2 base for generative detectors (DiffusionDet and DeFloMat). The Image Encoder for all generative models is a ResNet50 backbone pre-trained on ImageNet-1K. The Detection Decoder is a multi-stage transformer head. During training, we used the AdamW optimizer with a batch size of 24 for generative models. DeFloMat utilized a learning rate of $6 \times 10^{-5}$ and required 406 epochs ($\sim 136\text{k}$ iterations) to fully converge. Consistent with prior art, no data augmentation was applied during training due to the domain-specific nature of the clinical data.

\paragraph{Baselines and Ablation Study.}
We compare DeFloMat against established baselines: the two-stage Mask R-CNN, the one-stage YOLOV4, and the direct state-of-the-art generative detector, DiffusionDet. For fair comparison, both DiffusionDet and DeFloMat used $\mathbf{120}$ initial proposals ($N_{prop}=120$) during inference. The generative models were tested across a range of sampling steps ($S=1$ to $4$) to assess efficiency. Our ablation study specifically isolates the contribution of the proposed Flow Matching Loss, $\mathcal{L}_{\text{flow}}$, by comparing the full model (DeFloMat B+F) against a variant trained only with bounding box regression losses (DeFloMat B). We also tune the ratio of $\lambda_{\text{flow}}/\lambda_{L1}$ to determine the optimal balance.

\paragraph{Evaluation Metrics.}
Evaluation is performed slice-wise, comparing predicted bounding boxes against ground truth. The primary metrics are Average Precision (AP) at various Intersection over Union (IoU) thresholds: $\mathbf{AP}_{10}$, $\mathbf{AP}_{30}$, and $\mathbf{AP}_{50}$. We also report the combined $\mathbf{AP}_{10:50}$ metric (averaged over $10\%$ to $50\%$ IoU in $10\%$ increments). Performance is further characterized by $\mathbf{Precision}$ and $\mathbf{Recall}$ at key thresholds. For post-processing, we apply Non-Maximum Suppression (NMS) with an IoU threshold of $10\%$ and a confidence filter of $50\%$. Only slices containing ground-truth annotations were included in the metric calculation.

\subsection{Results}

\paragraph{Performance Comparison.}
Table~\ref{tab:performance_comparison} summarizes the detection accuracy of DeFloMat (Ours) compared to YOLOv4, Mask R-CNN, and DiffusionDet across various confidence thresholds ($\tau_c$). For the generative models, results are shown at $N_{prop}=120$ proposals and the efficient $S=3$ steps.

DeFloMat consistently achieves the highest overall detection accuracy. At $\tau_c = 30\%$, DeFloMat reaches an $\mathbf{AP}_{10}$ of $\mathbf{54.37\%}$, significantly outperforming DiffusionDet ($\mathbf{49.13\%}$) and Mask R-CNN ($\mathbf{40.63\%}$). Furthermore, DeFloMat excels in detection sensitivity, delivering a $\mathbf{Recall}@10$ of $\mathbf{59.17\%}$, exceeding DiffusionDet ($\mathbf{51.75\%}$) by $\mathbf{7.42}$ percentage points. This high recall is crucial for medical screening, where minimizing false negatives (missed lesions) is paramount. Even at the stricter threshold $\tau_c = 50\%$, DeFloMat maintains its leading performance, confirming its robust localization quality across IoU requirements. This superiority highlights the effectiveness of the flow-based loss in stabilizing box predictions near the ground truth, particularly in the high-efficiency regime.

\begin{table*}[t!]
\centering
\small
\setlength{\tabcolsep}{8pt}
\caption{\textbf{Ablation Study on the Efficacy of Flow Matching Loss.} Performance comparison (measured at $N_{prop}=120$ and $S=3$ steps) with and without the proposed Flow Matching Loss ($\mathcal{L}_{\text{flow}}$). The inclusion of $\mathcal{L}_{\text{flow}}$ provides a significant boost, resulting in a $\mathbf{5.13}$ percentage point increase in $\text{AP}_{10:50}$ (from 38.19 to 43.32). This confirms that explicitly regularizing the model to follow the deterministic Optimal Transport path is essential for achieving superior and stable localization accuracy in the few-step regime.}
\label{tab:ablation1}
\begin{tabular}{l|ccc|ccccc}
\toprule
& \multicolumn{3}{c|}{\textbf{loss}}
& \multicolumn{5}{c}{ } \\  
\cmidrule(lr){2-4}
\textbf{Methods}
& \textbf{L1} & \textbf{GIoU} & \textbf{Flow}
& \textbf{AP$_{10:50}$} & \textbf{AP$_{30}$} & \textbf{AP$_{50}$}
& \textbf{Precision} & \textbf{Recall} \\
\midrule

DiffusionDet
& \checkmark & \checkmark &
& 30.43 & 31.65 & 24.08 & 41.53 & 31.35 \\
\midrule

\multirow{3}{*}{DeFloMat}
& \checkmark & \checkmark & 
& 38.19 & 39.73 & 28.48 & 46.51 & 40.11 \\


& \checkmark & \checkmark & \checkmark
& \textbf{43.32} & \textbf{44.81} & \textbf{31.16}
& \textbf{48.21} & \textbf{46.62} \\
\bottomrule

\end{tabular}
\end{table*}

\paragraph{Qualitative Assessment.}
Figure~\ref{fig:examples}  provides visual evidence of the models' performance on challenging MRE slices, where inflammatory lesions can be subtle or obscured.
The qualitative results underscore DeFloMat's enhanced localization fidelity. In challenging cases (e.g., Row 2 and 4), Mask R-CNN and DiffusionDet frequently produce False Negatives (FN, Purple boxes), failing to detect clear inflammation sites. In contrast, DeFloMat reliably converts these FNs into True Positives (TP, Green boxes) by providing tighter and more accurate bounding box predictions. This suggests that the deterministic velocity field learned via Flow Matching is highly effective at precisely directing the proposal centers towards the true lesion locations, a characteristic that is vital for accurate clinical reporting. DeFloMat’s ability to achieve such precise localization with only $S=3$ steps highlights its clinical utility for rapid diagnostic auxiliary systems.

A core contribution of DeFloMat is its efficiency, achieved by learning the direct Optimal Transport path. This is rigorously validated in Figure~\ref{fig:vs}.

\paragraph{Stability among Number of Proposals.}
The stability of generative detectors against varying proposal counts is analyzed in Figure~\ref{fig:box_vs_ap}. This figure plots $\mathbf{AP}_{10:50}$ as the number of initial proposals ($N_{prop}$) increases from 12 to 120.

DeFloMat (orange lines) consistently outperforms DiffusionDet (blue lines) across the entire range of $N_{prop}$. Critically, DeFloMat's performance gain is sustained even with a high number of proposals ($N_{prop}=120$), where its performance stabilizes at an $AP_{10:50}$ around $43\%$. Conversely, while DiffusionDet also improves with more proposals, it remains significantly lower, indicating that its stochastic process is less efficient at leveraging the larger initial pool of proposals within the limited $S=3$ or $S=4$ steps. This demonstrates DeFloMat's robustness and scalability in exploring a larger search space without compromising efficiency.

\paragraph{Efficiency in Sampling Steps.}
The most significant finding regarding computational efficiency is presented in Figure~\ref{fig:step_vs_ap}, which compares $\mathbf{AP}_{10:50}$ against the number of sampling steps ($S$).

DeFloMat exhibits extraordinary convergence speed. Its performance rapidly increases from $S=1$ to $S=3$, where it effectively saturates at an $\mathbf{AP}_{10:50}$ of $\mathbf{43.32\%}$. This achievement stands in stark contrast to DiffusionDet, which requires many more steps to even approach its modest peak performance. At $S=4$, DeFloMat's performance is $\mathbf{1.4\times}$ higher than DiffusionDet's result, definitively confirming that the deterministic integration of the flow vector field resolves the latency bottleneck. The single-step ($S=1$) performance of DeFloMat is already highly competitive with DiffusionDet's multi-step results, validating the powerful effect of learning the direct Optimal Transport path for instantaneous and high-quality bounding box prediction.

\begin{table}[t]
\centering
\small
\setlength{\tabcolsep}{8pt} 
\caption{\textbf{Ablation Study on the Flow Loss Weight Ratio.} Impact of balancing the Flow Matching Loss ($\lambda_{\text{flow}}$) against the L1 Regression Loss ($\lambda_{L1}$) on overall detection performance (at $N_{prop}=120$ and $S=3$ steps). The optimal ratio of $\lambda_{\text{flow}} / \lambda_{L1} = \mathbf{0.1}$ yields the best results ($\text{AP}_{10:50}=43.32$), demonstrating that $\mathcal{L}_{\text{flow}}$ acts as a powerful regularization term. Higher weights lead to performance degradation, highlighting the necessity of carefully balancing flow dynamics with geometric localization objectives.}
\label{tab:ablation2}

\begin{tabular}{c|ccccc}
\toprule
\textbf{$\frac{\lambda_{flow}}{\lambda_{L1}}$} 
& \textbf{AP$_{10:50}$}
& \textbf{AP$_{30}$}
& \textbf{AP$_{50}$}
& \textbf{Precision}
& \textbf{Recall} \\
\midrule

1.0 
& 39.06 & 40.52 & 28.75 & 47.19 & 41.35 \\

0.5 
& 38.92 & 40.28 & 28.86 & 46.56 & 41.42 \\

0.2 
& 39.20 & 40.90 & 27.79 & 44.77 & 41.57 \\

0.1 
& \textbf{43.32} & \textbf{44.81} & \textbf{31.16} & \textbf{48.21} & \textbf{46.62} \\

\bottomrule
\end{tabular}
\end{table}

\subsection{Ablation Study}

\paragraph{Efficacy of the Flow Matching Loss.}
We first investigate the fundamental contribution of the deterministic flow regularization term. Table~\ref{tab:ablation1}  compares three scenarios, all evaluated at the efficient setting of $N_{prop}=120$ and $S=3$ steps: a DiffusionDet baseline, DeFloMat trained without $\mathcal{L}_{\text{flow}}$ (using only $\mathcal{L}_{L1}$ and $\mathcal{L}_{\text{GloU}}$), and the full DeFloMat model.

The results clearly demonstrate that the inclusion of $\mathcal{L}_{\text{flow}}$ provides a significant performance gain. Compared to the internal baseline (DeFloMat without $\mathcal{L}_{\text{flow}}$), the full model exhibits a substantial boost in $\mathbf{AP}_{10:50}$ from $38.19\%$ to $\mathbf{43.32\%}$, representing a $\mathbf{5.13}$ percentage point improvement. This gain confirms our core hypothesis: explicitly training the model to match the deterministic Optimal Transport velocity field is superior to relying solely on the sparse geometric regression losses ($\mathcal{L}_{L1}, \mathcal{L}_{\text{GloU}}$) applied at the end of the trajectory. The flow loss acts as an effective continuous supervisor, forcing the detection decoder ($\mathbf{v}_{\theta}$) to learn the most efficient direction of movement at every continuous time $t \in [0, 1]$.

\paragraph{Tuning of the Flow Loss Weight ($\lambda_{\text{flow}}$).}

The weight of the flow loss, $\lambda_{\text{flow}}$, is critical for balancing the deterministic flow dynamics with the precise geometric demands of bounding box regression. We tune the ratio $\lambda_{\text{flow}} / \lambda_{L1}$ while keeping $\lambda_{L1}$ constant (Table~\ref{tab:ablation2}).

The optimal performance is achieved at a relatively small ratio, $\mathbf{\lambda_{\text{flow}} / \lambda_{L1}} = \mathbf{0.1}$, resulting in the peak $\mathbf{AP}_{10:50}=43.32\%$. As the flow weight ratio increases (e.g., to $0.5$ or $1.0$), the performance degrades. At a ratio of $1.0$, the $\mathbf{AP}_{10:50}$ drops to $39.06\%$. This degradation occurs because an overly dominant flow loss causes the model to prioritize matching the direction of the straight-line path over satisfying the fine-grained geometric constraints necessary for accurate bounding box classification and coordinate prediction at the final step ($t=1$). The best result at $0.1$ validates $\mathcal{L}_{\text{flow}}$'s role as a powerful, yet delicate, regularization term that effectively enforces the desired deterministic dynamics without compromising localization precision.

\section{Conclusion}
\label{sec:discussion}

In this work, we introduced DeFloMat (Detection with Flow Matching), a novel framework that successfully integrates Conditional Flow Matching into the generative object detection paradigm to address the critical latency bottleneck of DiffusionDet. By replacing the unstable and slow stochastic diffusion process with the learning of a direct, deterministic vector field along an Optimal Transport path, DeFloMat achieved a major leap in efficiency and stability. Our extensive experiments on a clinical MRE dataset for Crohn's disease inflammation demonstrated state-of-the-art accuracy, specifically achieving an $\mathbf{AP}_{10:50}$ of $\mathbf{43.32\%}$ in only $S=3$ inference steps, which is a $\mathbf{1.4\times}$ performance improvement over DiffusionDet's maximum score. This remarkable convergence speed validates Flow Matching's potential in structured prediction tasks, positioning DeFloMat as a fast and reliable generative detection model.

\section*{Acknowledgments}
This work was supported by the Artificial Intelligence Industrial Convergence Cluster Development Project funded by the Ministry of Science and ICT (MSIT, Korea) \& Gwangju Metropolitan City, and by the National Research Foundation of Korea (NRF) grant funded by the Korea government (MSIT) (RS-2024-00336063, RS-2025-00520184).

{
    \small
    \bibliographystyle{ieeenat_fullname}
    \bibliography{main}

@String(CVPR= {IEEE Conf. Comput. Vis. Pattern Recog.})

@String(ICCV= {Int. Conf. Comput. Vis.})

@String(ECCV= {Eur. Conf. Comput. Vis.})

@String(ICLR = {Int. Conf. Learn. Represent.})

@String(CVPR  = {CVPR})

@String(ICCV  = {ICCV})

@String(ECCV  = {ECCV})

@String(ICLR  = {ICLR})

@article{Tsai2019MRE,
  author  = {Tsai, Richard and Mintz, Aaron and Lin, Michael and Mhlanga, Joyce and Chiplunker, Adeeti and Salter, Amber and Ciorba, Matthew and Deepak, Parakkal and Fowler, Kathryn},
  title   = {Magnetic resonance enterography features of small bowel Crohn's disease activity: an inter-rater reliability study of small bowel active inflammation in clinical practice setting},
  journal = {BMC Gastroenterology},
  volume  = {19},
  number  = {105},
  pages   = {1--11},
  year    = {2019},
  doi     = {10.1186/s12876-019-0985-0},
  url     = {https://pmc.ncbi.nlm.nih.gov/articles/PMC6636275/}
}

@article{Taylor2018METRIC,
  author  = {Taylor, Stuart A. and Mallett, Susan and Bhatnagar, Gauraang and Baldwin-Cleland, Rachel and Bloom, Stuart and Gupta, Arun and Hamlin, Peter J. and Hart, Ailsa L. and Higginson, Antony and Jacobs, Ilan and McCartney, Sara and Miles, Anne and Murray, Charles D. and Plumb, Andrew A. and Pollok, Richard C. and Punwani, Shonit and Quinn, Laura and Rodriguez-Justo, Manuel and Shabir, Zainib and Slater, Andrew and Tolan, Damian and Travis, Simon and Windsor, Alastair and Wylie, Peter and Zealley, Ian and Halligan, Steve and the METRIC study investigators},
  title   = {Diagnostic accuracy of magnetic resonance enterography and small bowel ultrasound for the extent and activity of newly diagnosed and relapsed Crohn's disease ({METRIC}): a multicentre trial},
  journal = {The Lancet Gastroenterology \& Hepatology},
  volume  = {3},
  issue   = {8},
  pages   = {548--558},
  year    = {2018},
  doi     = {10.1016/S2468-1253(18)30161-4},
  url     = {https://pmc.ncbi.nlm.nih.gov/articles/PMC6278907/}
}

@inproceedings{Chen2023DiffusionDet_ICCV,
  author    = {Chen, Shoufa and Sun, Peize and Song, Yibing and Luo, Ping},
  title     = {DiffusionDet: Diffusion Model for Object Detection},
  booktitle = {Proceedings of the IEEE/CVF International Conference on Computer Vision (ICCV)},
  pages     = {19830--19843},
  year      = {2023},
  url       = {https://openaccess.thecvf.com/content/ICCV2023/papers/Chen\_DiffusionDet\_Diffusion\_Model\_for\_Object\_Detection\_ICCV\_2023\_paper.pdf}
}

@article{liu2022flow,
  author  = {Liu, Xingchao and Chen, Ricky T. Q. and Welling, Max},
  title   = {Flow Straight and Fast: Learning to Generate and Transfer Data with Rectified Flow},
  journal = {arXiv preprint arXiv:2211.04208},
  year    = {2022},
  arxiv   = {2211.04208},
  url     = {https://arxiv.org/abs/2211.04208}
}

@article{Lipman2022FlowMatching,
  author  = {Lipman, Yaron and Chen, Ricky T. Q. and Ben-Hamu, Heli and Nickel, Maximilian and Le, Matt},
  title   = {Flow Matching for Generative Modeling},
  journal = {arXiv preprint arXiv:2210.02747},
  year    = {2022},
  arxiv   = {2210.02747},
  url     = {https://arxiv.org/abs/2210.02747}
}

@inproceedings{He2017MaskRCNN,
  author    = {He, Kaiming and Gkioxari, Georgia and Dollár, Piotr and Girshick, Ross},
  title     = {Mask {R-CNN}},
  booktitle = {Proceedings of the IEEE International Conference on Computer Vision (ICCV)},
  pages     = {2961--2969},
  year      = {2017},
  arxiv     = {1703.06870},
  url       = {https://arxiv.org/abs/1703.06870}
}

@article{Bochkovskiy2020YOLOv4,
  author  = {Bochkovskiy, Alexey and Wang, Chien-Yao and Liao, Hong-Yuan Mark},
  title   = {{YOLOv4}: Optimal Speed and Accuracy of Object Detection},
  journal = {arXiv preprint arXiv:2004.10934},
  year    = {2020},
  arxiv   = {2004.10934},
  url     = {https://arxiv.org/abs/2004.10934}
}

@inproceedings{girshick2015fast,
  title={{Fast R-CNN} : Towards Real-Time Object Detection with Region Proposal Networks},
  author={Girshick, Ross},
  booktitle={Proceedings of the IEEE International Conference on Computer Vision (ICCV)},
  pages={1440--1448},
  year={2015},
  doi={10.1109/ICCV.2015.169}
}

@inproceedings{carion2020detr,
  author    = {Carion, Nicolas and Massa, Francisco and Synnaeve, Gabriel and Usunier, Nicolas and Kirillov, Alexander and Zagoruyko, Sergey},
  title     = {End-to-End Object Detection with Transformers},
  booktitle = {European Conference on Computer Vision (ECCV)},
  pages     = {213--229},
  year      = {2020}
}

@article{rimola2011maria,
  author = {Rimola, Jordi and Rodr{\'{\i}}guez, Susanna and Garc{\'{\i}}a-Bosch, Oriol and Ord{\'a}s, Ingrid and Ayala, Emilio and Aceituno, Manuel and Pellise, Maria and et al.},
  title = {Magnetic Resonance for Assessment of Disease Activity and Severity in Crohn’s Disease},
  journal = {Gut},
  volume = {60},
  number = {8},
  pages = {1117--1125},
  year = {2011},
  doi = {10.1136/gut.2010.224808}
}

@article{ecco2017guidelines,
  author = {Gomoll{\'o}n, Fernando and Dignass, Axel and Annese, Vito and Tilg, Herbert and Van Assche, Gert and Lindsay, Joe O. and Peyrin-Biroulet, Laurent and et al.},
  title = {3rd European Evidence-based Consensus on the Diagnosis and Management of Crohn’s Disease 2016},
  journal = {Journal of Crohn's and Colitis},
  volume = {11},
  number = {1},
  pages = {3--25},
  year = {2017},
  doi = {10.1093/ecco-jcc/jjw168}
}

@inproceedings{sun2021sparsercnn,
  title     = {Sparse R-CNN: End-to-End Object Detection with Learnable Proposals},
  author    = {Sun, Peize and Zhang, Yi and Jiang, Yue and Kong, Tao and Xu, Chunhua and Wei, Zhiqiang and Xu, Zijie and Li, Junjie and Jia, Jiaya},
  booktitle = {Proceedings of the IEEE/CVF Conference on Computer Vision and Pattern Recognition (CVPR)},
  pages     = {14454--14463},
  year      = {2021},
  doi       = {10.1109/CVPR46437.2021.01422}
}

@inproceedings{bodla2017softnms,
  author    = {Bodla, Navaneeth and Singh, Bharat and Chellappa, Rama and Davis, Larry S.},
  title     = {{Soft-NMS}-- Improving Object Detection With One Line of Code},
  booktitle = {Proceedings of the IEEE International Conference on Computer Vision (ICCV)},
  year      = {2017},
  pages     = {5561--5569},
  doi       = {10.1109/ICCV.2017.593}
}

@inproceedings{rezatofighi2019giou,
  title     = {Generalized Intersection over Union: A Metric and A Loss for Bounding Box Regression},
  author    = {Rezatofighi, Seyed and Tsoi, Anton and Gwak, JunYoung and Sadeghian, Amir and Reid, Ian and Savarese, Silvio},
  booktitle = {Proceedings of the IEEE/CVF Conference on Computer Vision and Pattern Recognition (CVPR)},
  pages     = {658--666},
  year      = {2019},
  doi       = {10.1109/CVPR.2019.00075}
}

@inproceedings{ho2020ddpm,
  title     = {Denoising Diffusion Probabilistic Models},
  author    = {Ho, Jonathan and Jain, Ajay and Abbeel, Pieter},
  booktitle = {Advances in Neural Information Processing Systems (NeurIPS)},
  volume    = {33},
  pages     = {6840--6851},
  year      = {2020}
}

@inproceedings{song2021ddim,
  title     = {Denoising Diffusion Implicit Models},
  author    = {Song, Jiaming and Meng, Chenlin and Ermon, Stefano},
  booktitle = {International Conference on Learning Representations (ICLR)},
  year      = {2021},
  url       = {https://arxiv.org/abs/2010.02502}
}

@article{deng2009imagenet,
  title     = {ImageNet: A Large-Scale Hierarchical Image Database},
  author    = {Deng, Jia and Dong, Wei and Socher, Richard and Li, Li-Jia and Li, Kai and Fei-Fei, Li},
  journal   = {Proceedings of the IEEE Conference on Computer Vision and Pattern Recognition (CVPR)},
  pages     = {248--255},
  year      = {2009},
  doi       = {10.1109/CVPR.2009.5206848}
}

@inproceedings{lin2017fpn,
  title     = {Feature Pyramid Networks for Object Detection},
  author    = {Lin, Tsung-Yi and Doll{\'a}r, Piotr and Girshick, Ross and He, Kaiming and Hariharan, Bharath and Belongie, Serge},
  booktitle = {Proceedings of the IEEE Conference on Computer Vision and Pattern Recognition (CVPR)},
  pages     = {936--944},
  year      = {2017},
  doi       = {10.1109/CVPR.2017.106}
}

@inproceedings{he2016resnet,
  title     = {Deep Residual Learning for Image Recognition},
  author    = {He, Kaiming and Zhang, Xiangyu and Ren, Shaoqing and Sun, Jian},
  booktitle = {Proceedings of the IEEE Conference on Computer Vision and Pattern Recognition (CVPR)},
  pages     = {770--778},
  year      = {2016},
  doi       = {10.1109/CVPR.2016.90}
}
}


\end{document}